\documentclass[fleqn,10pt]{wlscirep}
\usepackage[utf8]{inputenc}
\usepackage[T1]{fontenc}
\usepackage{lineno}
\usepackage{multirow}
 
\newcommand{\red}[1]{\textcolor[rgb]{0,0,0}{#1}} 

\title{RAID-Database: human Responses to Affine Image Distortions}

\author[1,*]{Paula Daudén-Oliver}
\author[2,3]{David Agost-Beltran}
\author[2,3]{Emilio Sansano-Sansano}
\author[2,3]{Raul Montoliu}
\author[1]{Valero Laparra}
\author[1]{Jesús Malo}
\author[2,4]{Marina Martínez-Garcia}

\affil[1]{Universitat de València, Image Processing Laboratory, Valencia, 46022, country}
\affil[2]{Universitat Jaume I, Castellón, 12071, Spain}
\affil[3]{Institute of New Imaging Technologies, \emph{Universitat Jaume I}, 12071 Castellón, Spain}

\affil[4]{Institut de Matemàtiques de Castelló, \textit{Universitat Jaume I}, Castell\'{o} 12071, Spain}

\affil[*]{corresponding author(s): Paula Daudén (paula.dauden@uv.es)}

\begin{abstract}
Image quality databases are used to train models for predicting subjective human perception. 
However, most existing databases focus on distortions commonly found in digital media and not in natural conditions. 
Affine transformations are particularly relevant to study, as they are among the most commonly encountered by human observers in everyday life. 
This Data Descriptor presents a set of human responses to suprathreshold affine image transforms (rotation, translation, scaling) and Gaussian noise as convenient reference to compare with previously existing image quality databases. 
The~responses were measured using well established psychophysics: the Maximum Likelihood Difference Scaling method.
The set contains responses to 864 distorted images.
The experiments involved 105 observers and more than 20000 comparisons of quadruples of images. The quality of the dataset is ensured because 
(a) it reproduces the classical Piéron's law, 
(b) observers obtain reasonable psychometric functions depending on the stimulation strength. 
(c) it reproduces classical absolute detection thresholds, and 
(d) it is consistent with conventional image quality databases but improves them according to Group-MAD experiments.

\end{abstract}
\begin{document}

\flushbottom
\maketitle

\thispagestyle{empty}


\section*{Background \& Summary}


Experimental data on subjective differences between images is very useful both in engineering and basic science.
On the one hand, the evaluation of subjective image quality is a critical task in image processing. 
Many perceptual metrics used for different tasks are designed and trained to correlate the subjective quality of distorted images with human perception~\cite{lpips}.%
On the other hand, subjective image distortion has also been used to fit human vision 
models/metrics ~\cite{Malo02, Perceptnet}.
In fact, correlation with subjective distortions has been used to evaluate the human nature of artificial vision systems,
and there is a debate about the quality of such databases to be used in these scientific applications  ~\cite{Martinez19}.
As a consequence, databases containing distorted images paired with human subjective quality scores are essential. 

Given the engineering origin of the most common databases, they focus on distortions found in digital environments rather than natural ones~\cite{kadid,pipal,tid2013,CID2013}. 
In contrast, the usual type of transformations in nature are due to changes in point of view, motion of objects and changes in illumination.
These are all affine transforms of the image: rotations, translations and scaling (affine in the spatial domain), or illumination changes (affine in the amplitude of the signal).
Contrast variation of noise which is usual in conventional databases~\cite{...}, and controlled illumination changes which is not that common, but has also been reported,
are extremely useful to fit low-level adaptation models.
Similarly, the response to geometric affine transforms could be equally useful to improve mid-level models of shape perception.
In any case, studying how perceptual metrics respond to these transformations is important to determine whether their behaviour aligns with human perception~\cite{nuria}.

%
%
%
%
%
%


This Data Descriptor presents a set of human responses to suprathreshold affine image transforms (rotation, translation, scaling) and Gaussian noise as convenient reference to compare with previously existing image quality databases.

One of the most used databases is TID2013 \cite{tid2013}, which comprises 25 reference images (24 natural and 1 synthetic), 24 types of distortion and 5 levels for each type of distortion. Examples of the distortions included are different types of noise, compression artifacts, contrast and color changes.
In our study, the same 24 natural images were used as reference images, with distortions including rotation, translation, scaling and Gaussian noise. Gaussian noise was chosen to have a common distortion between both databases and distortion levels were carefully chosen to ensure that the maximum value in TID2013 aligns with the maximum value in our database. 
This alignment enables scaling both datasets to be used in combination and proper comparison of their relative quality.

Visibility of distortions or response to distortions can be quantified in many ways.
In particular, conventional databases (e.g.~\cite{tid2013,kadid}) use the concept of Mean Opinion Score (MOS), which tries to measure to how an average observer would rate a distorted image regard the undistorted one. 
These values can be obtained with different methods such as numerical ratings in a specific range, adjectival descriptors, or pairwise comparisons of different images~\cite{ITU500}, each with their own advantages and disadvantages~\cite{Psychophysics}. %
In our study we have chosen a method used in classical psychophysics called Maximum Likelihood Difference Scaling Method (MLDS) described by Maloney and Yang \cite{Maloney} to obtain the responses to the distorted images. This method compares stimuli of the same reference image and type of distortion, generating a curve of values for each stimulus. In the curves each point corresponds to the response to a level of distortion. It also provides a sigma value, considered to represent internal noise, which is can be used to scale each curve.
Using this method we get the response curve in a wide range of stimulation, from zero (below threshold) to values well above the threshold, thus generalizing classical measurements on detection thresholds.
Additionally, while performing MLDS reaction times were recorded.

The quality of the reported responses are validated in three ways: (1) they reproduce the classical detection thresholds in the low-stimulation limit, (2) they reproduce the classical Pieron's law (an its relation with the sensitivity of the observer, or slope of the response curve), and (3) the responses to white noise are consistent with conventional image quality databases, but interestingly, separate Group-MAD experiments show that our database better describes human behavior. 

For these reasons the proposed set of experimental data constitutes an interesting extension to conventional subjective image distoriton databases and it is fully compatible with the existing MOS results.






\section*{Methods}

\subsection*{Images}
The database contains a total of 888 images, including 24 reference images obtained from the  \href{http://r0k.us/graphics/kodak/}{Kodak Lossless True Color Image Suite}. and 864 images distorted with four types of distortions: rotation, translation, scaling, and Gaussian noise. Each distortion has nine incremental levels apart from the reference. As shown in Table \ref{tab:tabla_umbrales}, these distortion levels were selected to ensure that the range of human visual threshold falls within the first levels for each type of distortion. For rotation, the increment is 2 degrees per level, reaching a maximum rotation of 18 degrees. For translation, each step increases by 0.07 degrees (visual angle) horizontally. Scaling is applied in 1\% increments, reaching a maximum of 9\% of the image size. Finally, Gaussian noise is added in steps of 0.0009, ranging from 0 to a maximum of 0.008 in variance. The maximum level of Gaussian noise was chosen to closely match the maximum noise level in the TID2013 database, allowing the results to be comparable and allowing both data sets to be used together.

\begin{table}[!h]

\centering
\resizebox{13cm}{!} {
\begin{tabular}{|c|c|c|c|}
\cline{1-4} 
\textbf{Affine Transformation} & \textbf{Human Threshold} & \textbf{Step} & \textbf{Maximum} \\ \hline
\multicolumn{1}{|c|}{Rotation} & 3.6 degrees \cite{nuria} & 2 degrees & 18 degrees \\ \hline
\multicolumn{1}{|c|}{Translation} & 0.23 degrees \cite{nuria} & 0.07 degrees & 0.63 degrees \\ \hline
\multicolumn{1}{|c|}{Scale} & 1.026\% \cite{nuria} & 1\% & 9\%  \\ \hline
\multicolumn{1}{|c|}{Gaussian Noise} & - & 0.0009 & 0.008 \\ \hline
\end{tabular}
}
\caption{Steps and maximum values for the different transformations and human visibility thresholds.}
\label{tab:tabla_umbrales}
\end{table}

\begin{figure}[!h]
\centering
			\includegraphics[width=0.8\linewidth]{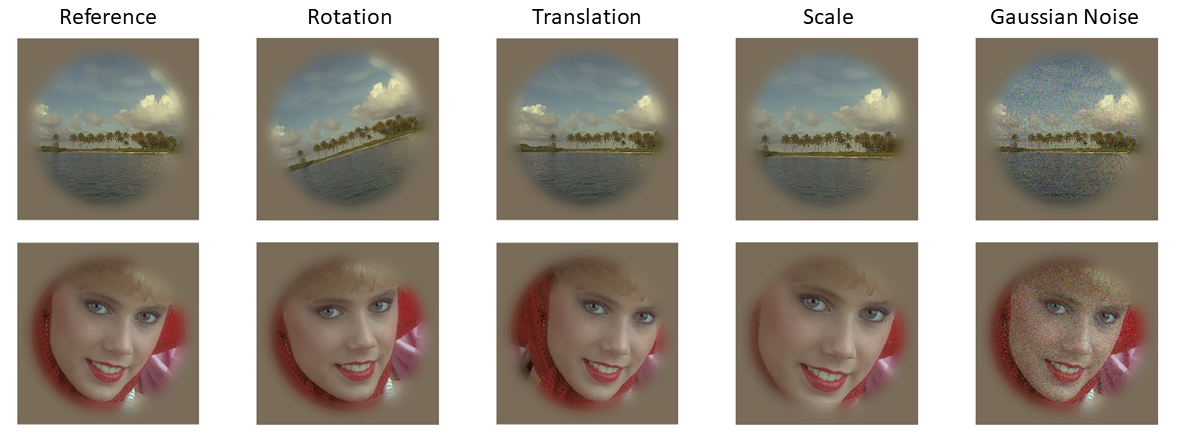}	
\caption{Example of the transformation of two reference images with the maximum level of each distortion. }
\label{fig:figura_img}
\end{figure}

\noindent The size of the images is 454x454x3. The images were cropped in a circular shape with smooth edges to prevent the observer from using the edges to perform the task, since in this type of transformation, it was inevitable that some objects appeared and disappeared from the image when a transformation was applied. 
The background color is the average color of the images. Figure \ref{fig:figura_img} shows an example of two reference images distorted with the maximum level of each transformation.


\subsection*{Experimental measurement of the responses}
\label{sec:exp_meas}

\begin{figure}[!h]
\centering
			\includegraphics[width=0.8\linewidth]{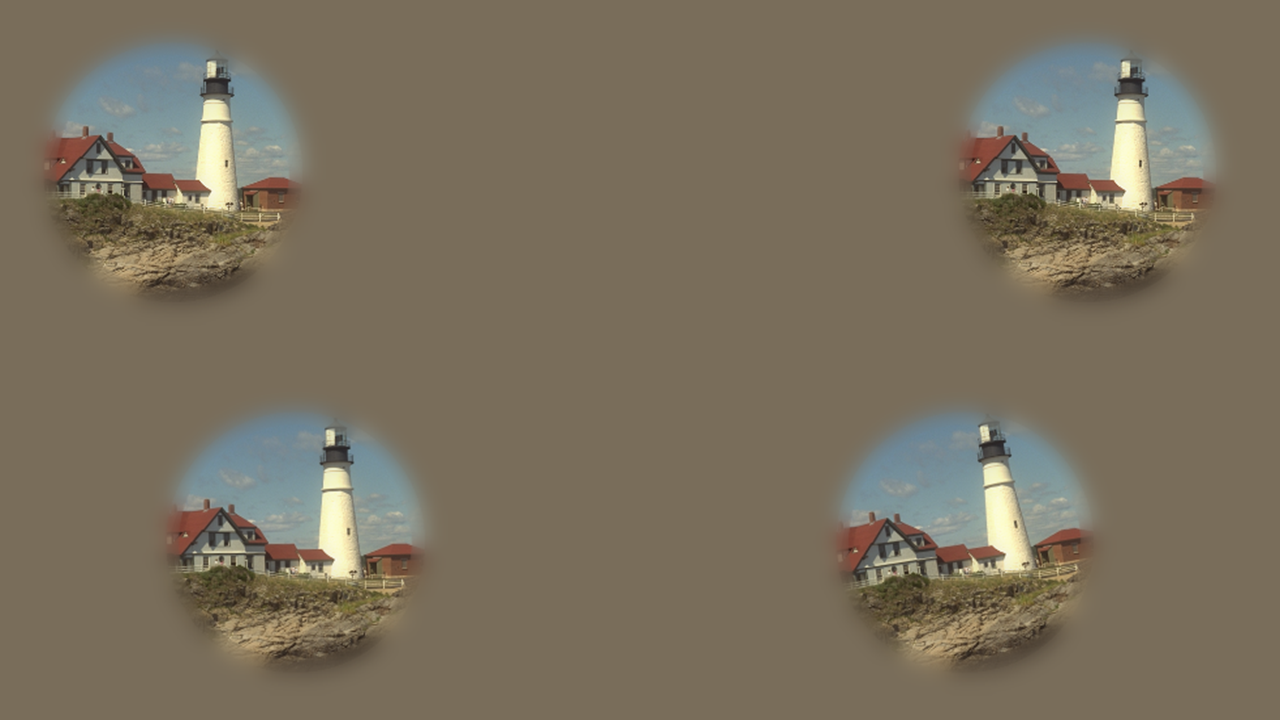}	
\caption{Example of a trial presented to the observer. }
\label{fig:ejemplo_software}
\end{figure}

To obtain the responses for each distorted image, The Maximum Likelihood difference scaling method (MLDS) described by Maloney and Yang\cite{Maloney} was used. This method gets a response curve for each reference image and distortion. This means that only images with the same distortion and reference are compared with each other, resulting in a total of 96 different curves (see Figure \ref{fig:cuatro_imagenes}). 
In each trial, four images were presented simultaneously: two on the left and two on the right (see figure \ref{fig:ejemplo_software}). Observers had to select which pair (left or right) had the largest perceptual difference. 210 quadruples were selected to measure each curve. 
The 10 stimuli were combined in a way that the two images in one pair $(x_1, x_2)$ had a lower distortion level than the two images in the other pair $(x_3, x_4)$. Also, for a quadruple $(x_1, x_2)$  and $(x_3, x_4)$, the difference $||x_1 - x_2| - |x_3 - x_4||$ \red{(call it $\Delta_{\textrm{\emph{difference}}})$} cannot be larger than 7. This decision was made to avoid combinations where the answer is too obvious in order to make the experiment shorter without removing relevant answers. A total of 105 observers, aged 18 to 60 years, participated in these measurements. Observers visited our laboratory and did the measures on the same monitor under controlled conditions as low external illumination and 80 cm of distance to the screen. Each image was presented on the screen with size 10 cm of diameter, which corresponds to 7.125 vision degrees. Each participant provided informed consent prior to participation and received a gift card as compensation. 

To prevent observer fatigue, all comparisons from the same distortion were mixed and presented randomly in blocks of 48 trials. With this, one observer did 4 blocks (one for each distortion) with 48 trials each, seeing different reference images across blocks. The maximum time they had to answer each trial was 1 minute, but we advised them to answer faster, about 10 seconds per trial (the actual time for each trial is provided in the database). After each block, observers could rest for a few minutes. The duration of the complete experiment was approximately 40 minutes (10 per block including resting time). As shown in Figure \ref{fig:ejemplo_software}, the pairs are presented in such a way that they are not aligned to prevent the observer from getting clues about horizontal displacements by creating imaginary lines between the images.

For the curve fitting, this method estimates weights that predict the observer’s judgment, except for the first and last parameters, which are fixed at 0 and 1, respectively. The method also estimates the internal noise, which is used to scale the curves. 

While conducting the MLDS experiments the reaction time was recorded because (1)~it is useful to assess the quality of the responses if they are consistent with the classical Pieron's law, 
and (2)~this behavior may be related to the sensitivity of the observers (or slope of the response curves).

\subsection*{Responses}

\begin{figure}[htbp]
    \centering
    \begin{minipage}[c]{7cm}
    \includegraphics[width=1\linewidth]{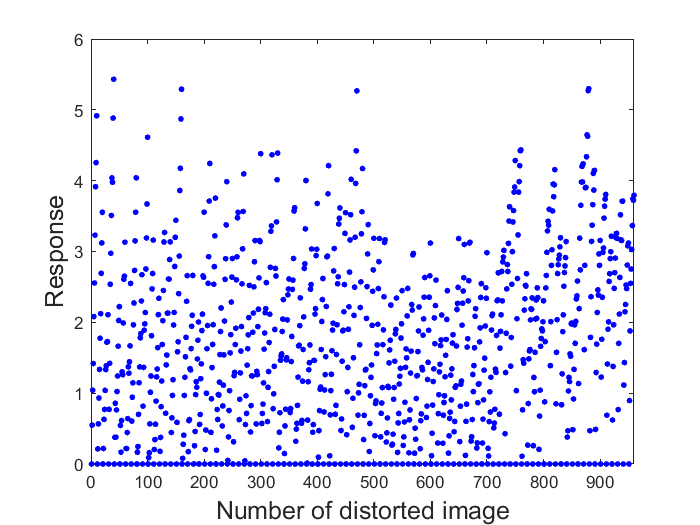} 
    \end{minipage}
    \begin{minipage}[c]{10cm}
    \begin{tabular}{cc}
    \includegraphics[width=0.5\textwidth]{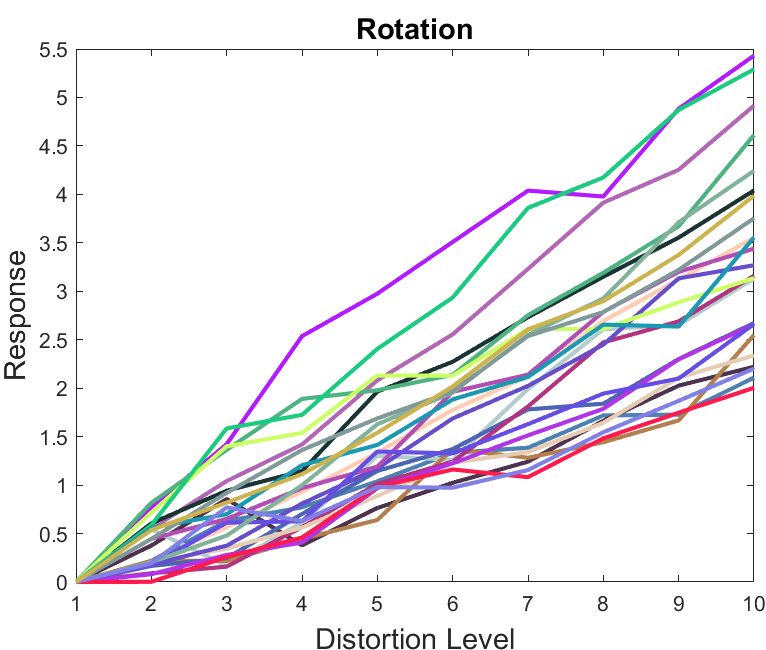} & 
    \includegraphics[width=0.48\textwidth]{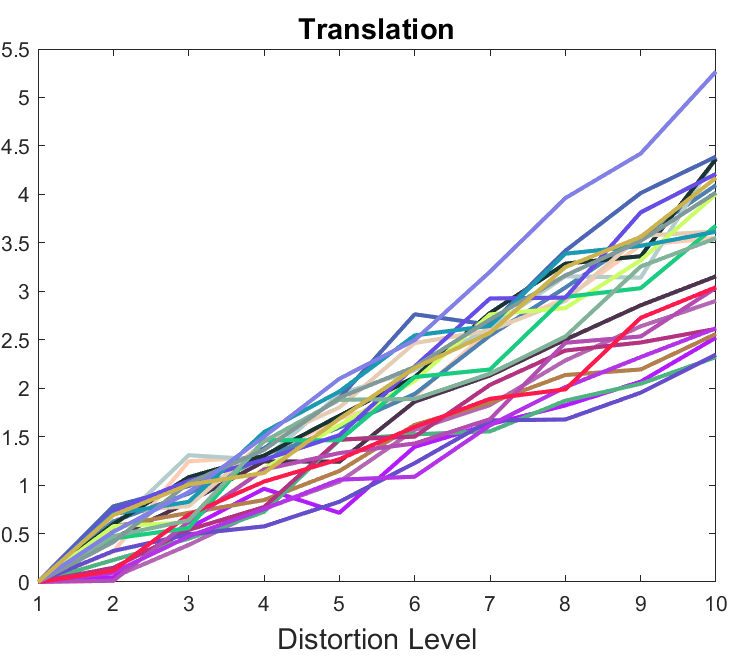} \\
    \includegraphics[width=0.48\textwidth]{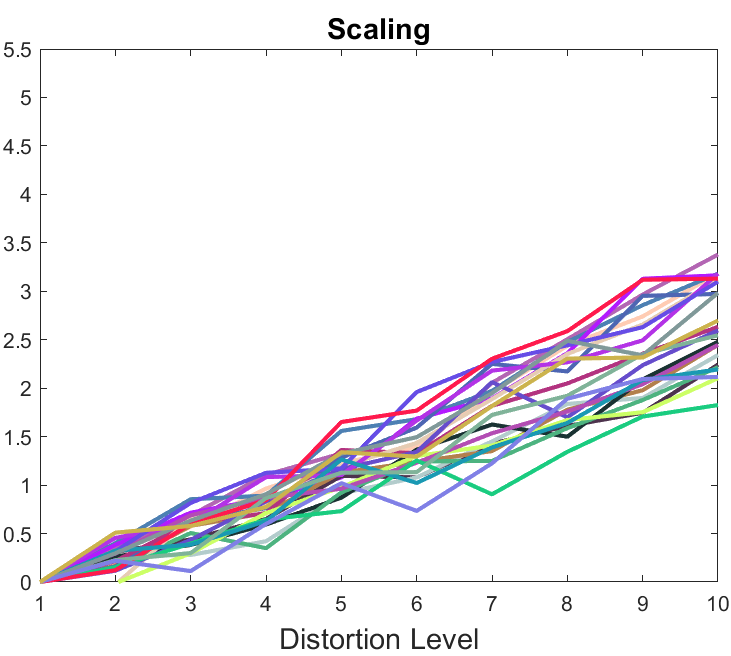} &
    \includegraphics[width=0.48\textwidth]{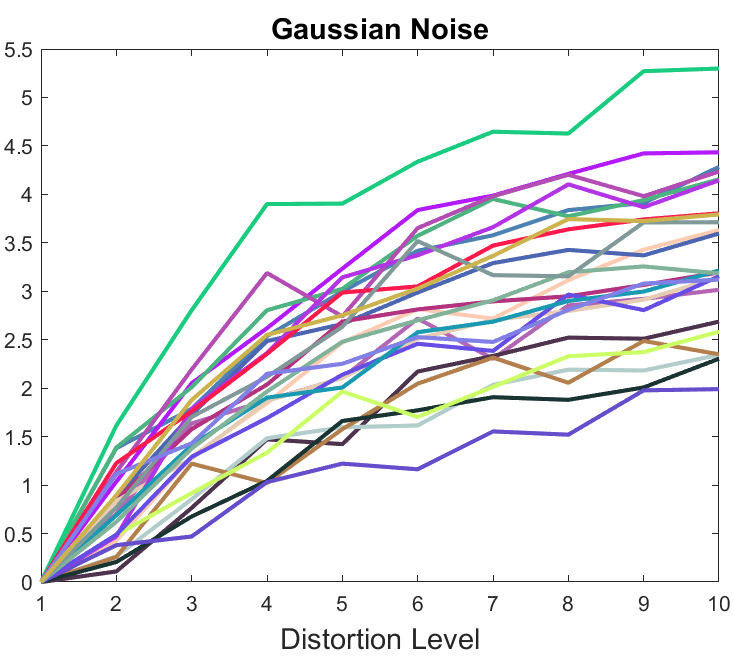}
    \end{tabular}
    \end{minipage}
    \caption{MLDS responses for all distorted images. Left: all the images with its corresponding response. Right: Each of the four plots corresponds to a particular distortion, each line corresponds to the response to an image with different levels of the distortion.}
    \label{fig:cuatro_imagenes}
\end{figure}

Figure  \ref{fig:cuatro_imagenes} illustrates all the response curves for different reference images and separated in distortions. All curves show an increase in the response as the distortion level increases. However, can be visually perceived that the increase for rotation, translation and scaling follow a more linear trend, whereas Gaussian noise exhibits a more saturating behavior as expected \cite{Daly}. 

\section*{Data Records}



\begin{figure}[!h]
\centering
			\includegraphics[width=0.8\linewidth]{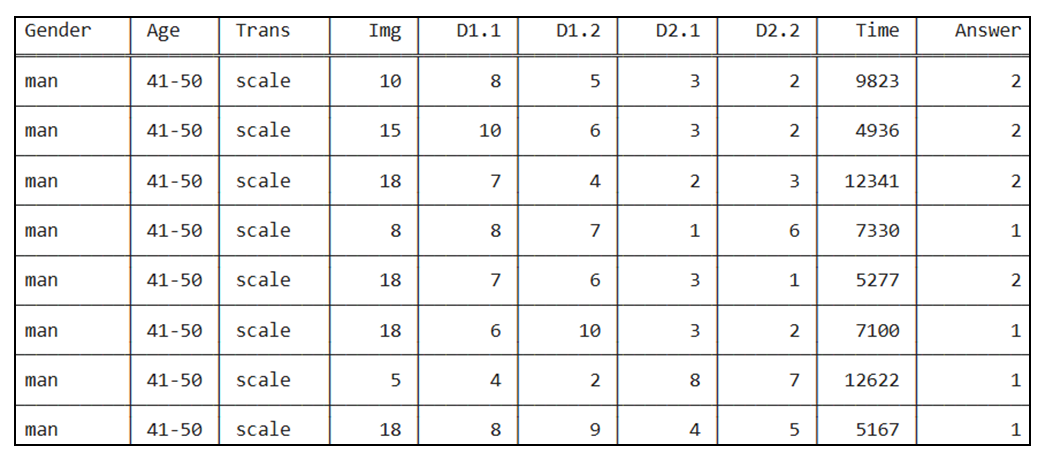}	
\caption{{\bf Raw data}: Example of the database of human recordings. Only the first rows are shown. See text for details.}
\label{fig:DDBB}
\end{figure}

\begin{figure}[!h]
\centering
			\includegraphics[width=0.5\linewidth]{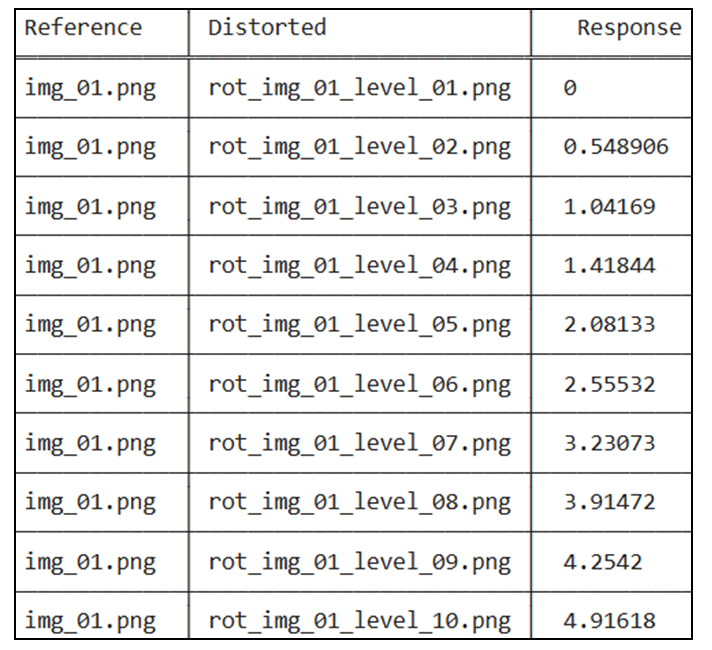}	
\caption{{\bf MLDS data}: Example of the database of MLDS responses computed from the raw data. Only the first rows are shown. See text for details.}
\label{fig:DDBB_curvas}
\end{figure}

The dataset collected in this study is available for download, see  Code availability section. We provide different two versions to download the dataset:
\begin{itemize}
    \item Raw data: are the data collected in the experiments from the human observers following the description in 'Experimental measurement' section. Figure \ref{fig:DDBB} shows an example of the database. Each row is a decision made by a human. Each decision has 10 features: two that depend on on the observer ('Gender', and 'Age'), the distortion applied ('Trans'), the original image to which the distortion is applied ('Img'), the four different distortion levels to be compared (two for one pair ['D1.1', 'D1.2'], and two for the other pair ['D2.1', 'D.2.2']), the response time in milliseconds ('Time'), and the selected pairs with larger difference ('Answer').   
    \item MLDS data: are the data that contain the computed MLDS responses. This data has been obtained by processing the \emph{raw data} using the MLDS method. Note that the MLDS responses are provided divided by the MLDS standard deviation, therefore the standard deviation for all the provided curves is 1. This makes the curves comparable among them. Example of the computed curves are shown in fig. \ref{fig:tres_imagenes} and fig. \ref{fig:curva_val}. The data format can be found in Figure \ref{fig:DDBB_curvas}. Each row corresponds to an MLDS response point. It has three columns: the first one is the name of the original image ('Reference'), the second column is the distorted image ('Distorted', the name makes reference to the original image, the type of distortion, and the amount of distortion), and the MLDS response ('Response').  
\end{itemize}

\section*{Technical Validation}
\label{sec:Tech_val}



Different technical validations have been performed. On the one hand we certified the Piéron's law directly on the raw data recordings. After computing the MLDS responses from the human responses, we certified that  the intra-image, the inter-images and the global responses are consistent with the literature and the expectations. As a third level validation, we certified the consistency of our results with TID2013 database. Besides, we perform a Group MAD competition\cite{gmad} against TID2013 to analyze the most critical differences.   

\begin{figure}[!h]
\centering
\begin{tabular}{cc}
\includegraphics[width=0.5\linewidth]{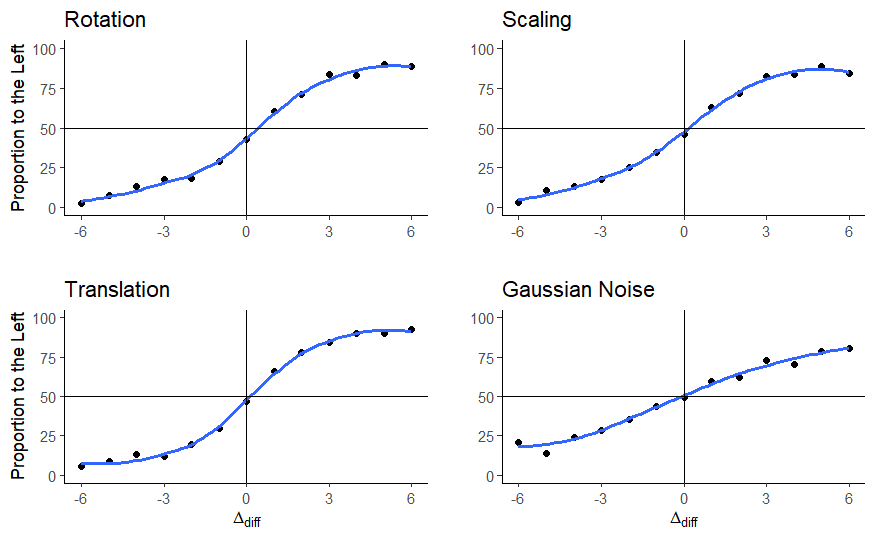} & 

\includegraphics[width=0.45\linewidth]{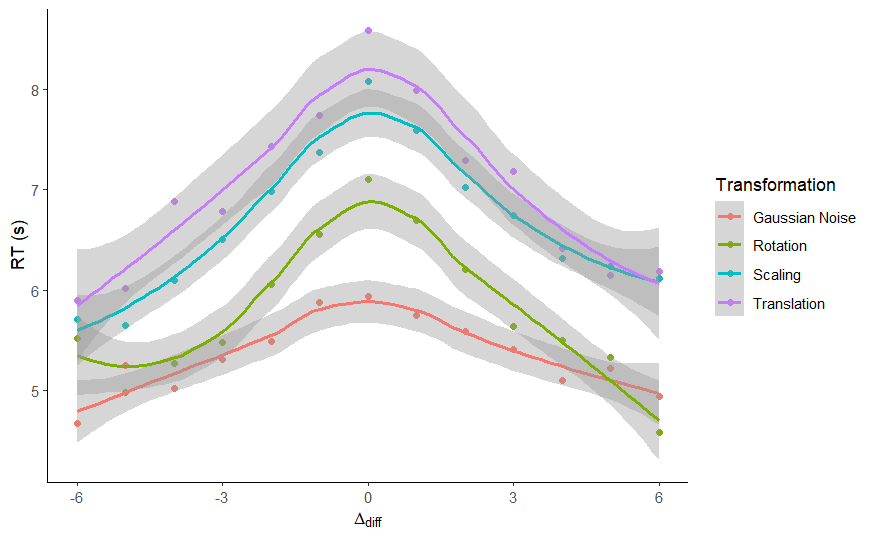}	
\end{tabular}
\caption{Left panel, the average probability to select the “left” option for all the subjects as a function of the  $\Delta_{diff}$ (the difference $||x_1 - x_2| - |x_3 - x_4||$ for the quadruplets). Right panel: Average Reaction time in seconds (RT) for for all the subjects as a function of the $\Delta_{diff}$ for all the distortion .
}
\label{fig:RT_all}
\end{figure}

{\bf The correlation of reaction time} (RT) and task difficulty has been shown by many experimental, [15–17] and explained by theoretical studies [18, 19]. The form of Piéron's Law, applied for each transformations, reflects a power-law relationship between stimulus degree of distortion and reaction time (RT) Fig.\ref{fig:RT_all} left.  Fig.\ref{fig:RT_all} emphasizes that as the distortion becomes stronger, the brain processes it faster, resulting in shorter reaction times. This is thought to arise from the brain’s neural efficiency in encoding higher-intensity signals. 
As shown in Fig.\ref{fig:RT_all}-Right, RT decreases as a function of the ease of the trial bigger $\Delta_{diff}$. This V inverted-shaped pattern has been previously associated to decision difficulty [5, 7, 20–22]. Therefore, and since we do not have a direct measure of difficulty for the different transformations,

\begin{figure}[htbp]
    \centering
    \includegraphics[width=1\textwidth]{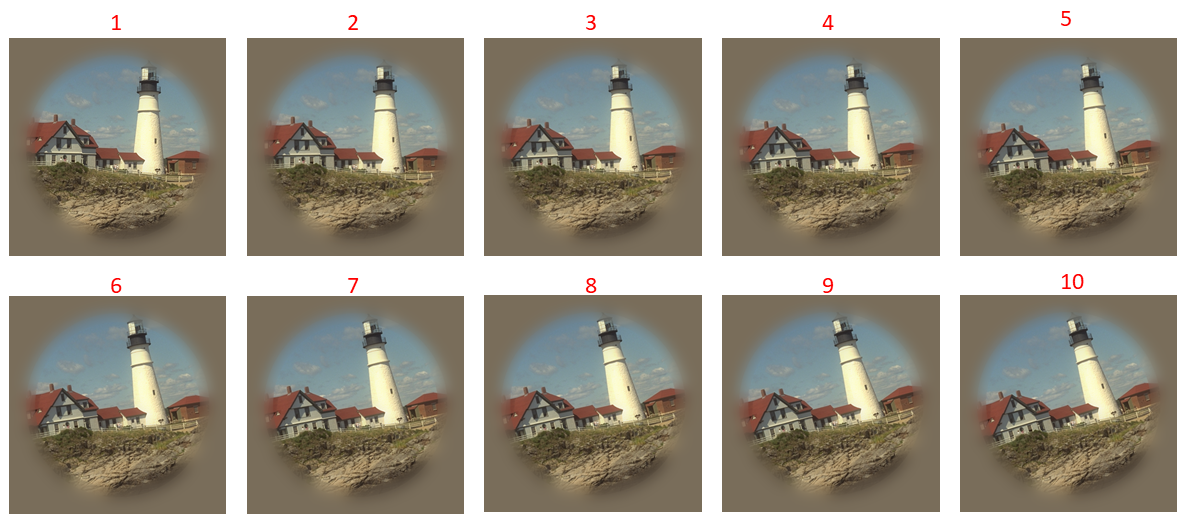}
    \includegraphics[width=1\textwidth]{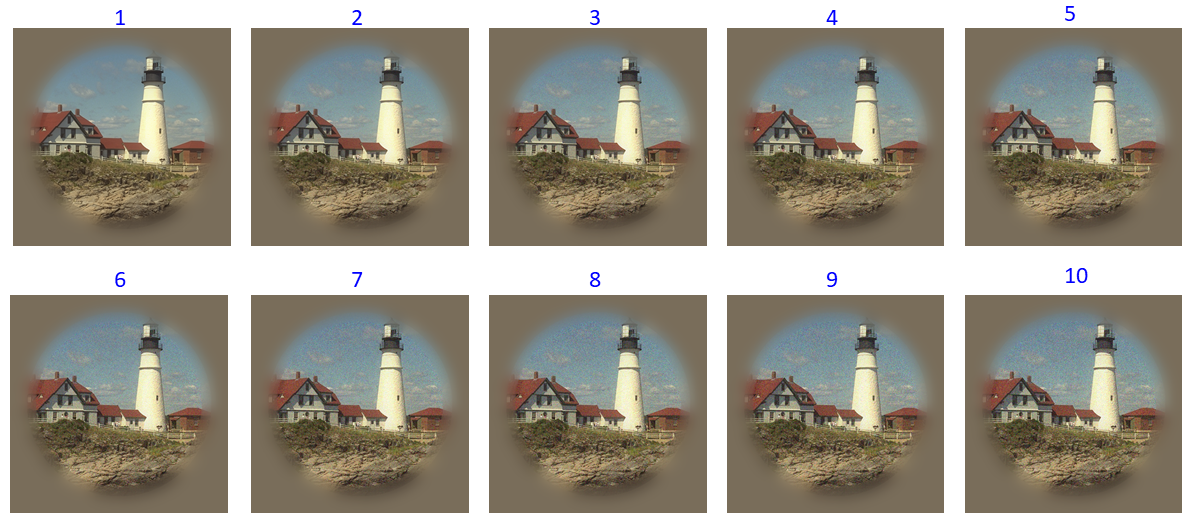}
    \includegraphics[width=1\textwidth]{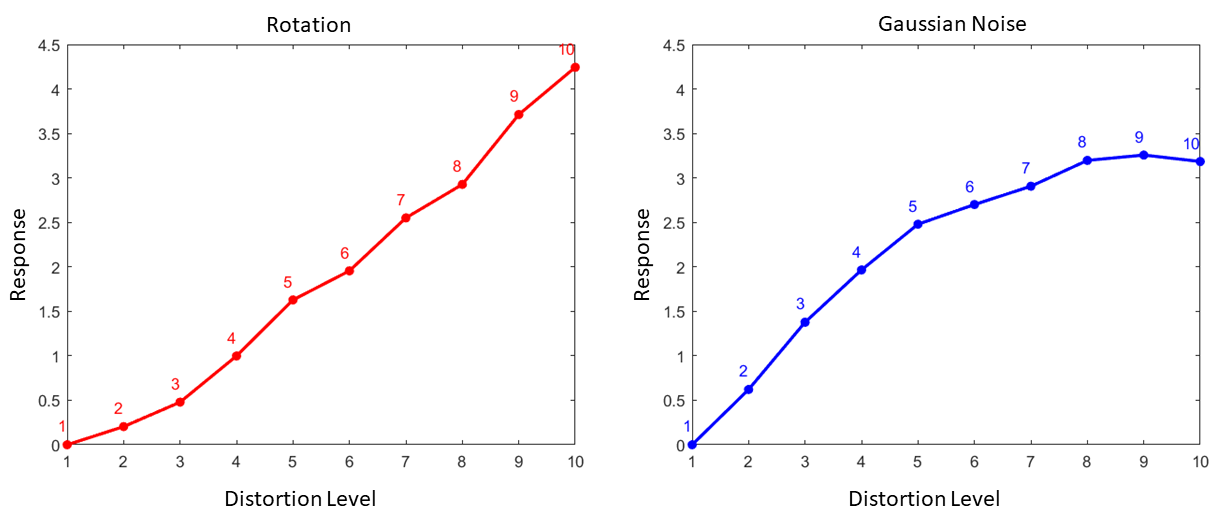}
    \caption{For one reference image, responses to rotation and Gaussian noise. Top: All levels of rotation (red), and Gaussian noise (blue). Bottom: Response curves to rotation (red) and Gaussian noise (blue). }
    \label{fig:tres_imagenes}
\end{figure}

{\bf Intra-image responses} are consistent with the expectations. Figure \ref{fig:tres_imagenes} shows a visual example of the MLDS responses computed from our experiments. In the case of rotation distortion the difference between each pair of rotated images is uniform (i.e. difference between images 2 vs 3, and 9 vs 10 are perceived equal), while in the case of noise the effect is not uniform (is perceived higher between images 2 vs 3, than for images 9 vs 10). 

\begin{figure}[htbp]
    \centering
    \includegraphics[width=0.8\textwidth]{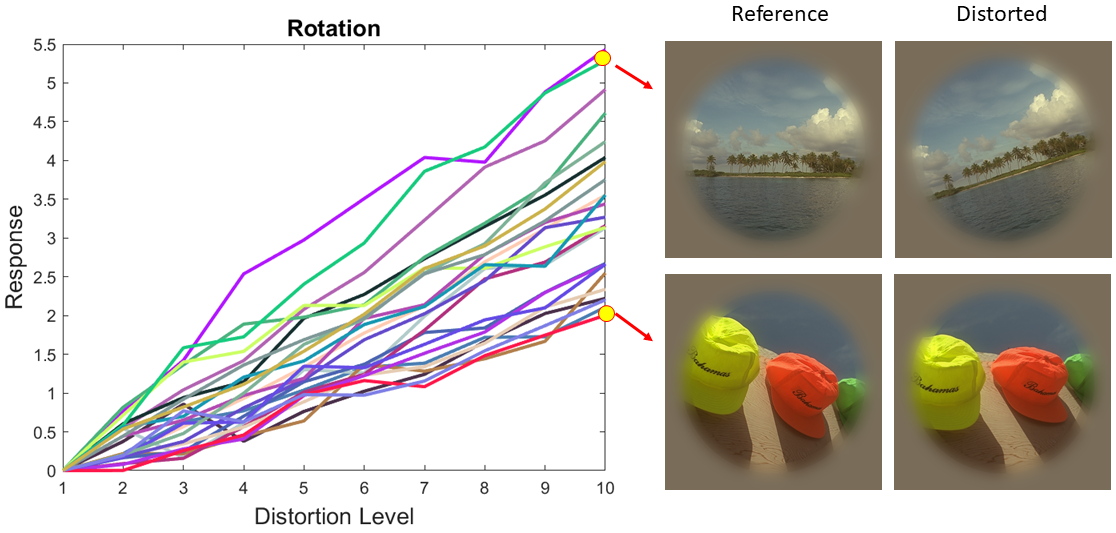}
    \includegraphics[width=0.8\textwidth]{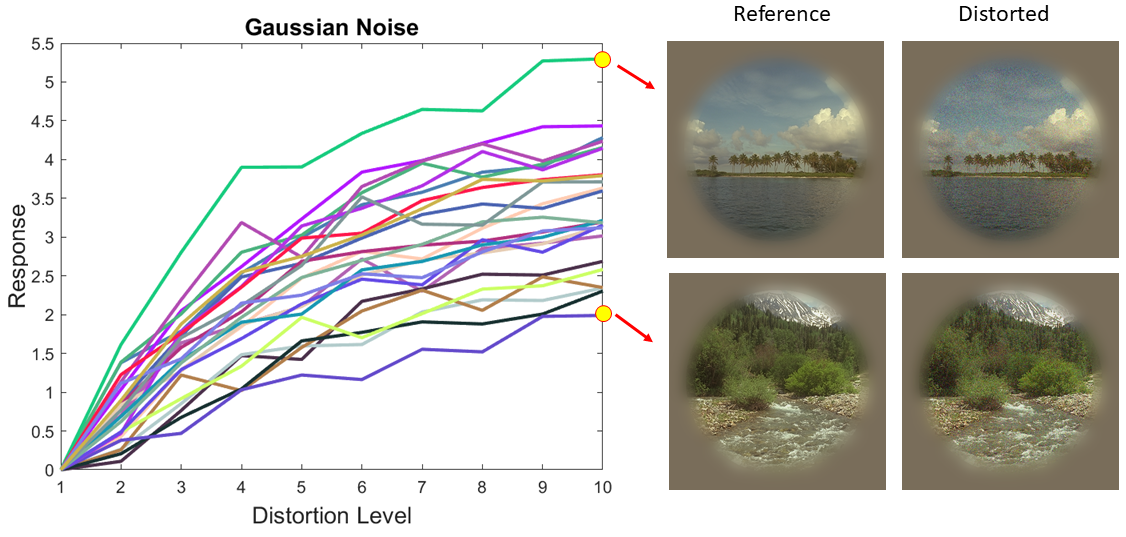}
    \caption{Responses of different reference images to the same distortion type and level. It can be observed that higher response values correspond to greater distortion visibility.} 
\label{fig:curva_val}
\end{figure}

{\bf Inter-images responses} are consistent with the expected behavior. Figure \ref{fig:curva_val} shows how for images where the same amount of rotation or gaussian noise is perceived differently, the MLDS responses have different value. In particular rotating an image with a marked horizon is highly perceived while the rotation of an image without horizontal features is less perceived. On the other hand, adding gaussian noise to an image with flat regions is more visible than in an image without low frequencies.      

{\bf Reproduction of classical Absolute Detection Thresholds.} 
Another proof of the quality of the experimental data in the presented set is that in the low stimulation range (for small distortions) one can derive the absolute detection threshold. 
The results are consistent with the classical literature on the issue.

\begin{figure}[!h]
\centering
			\includegraphics[width=1.0\linewidth]{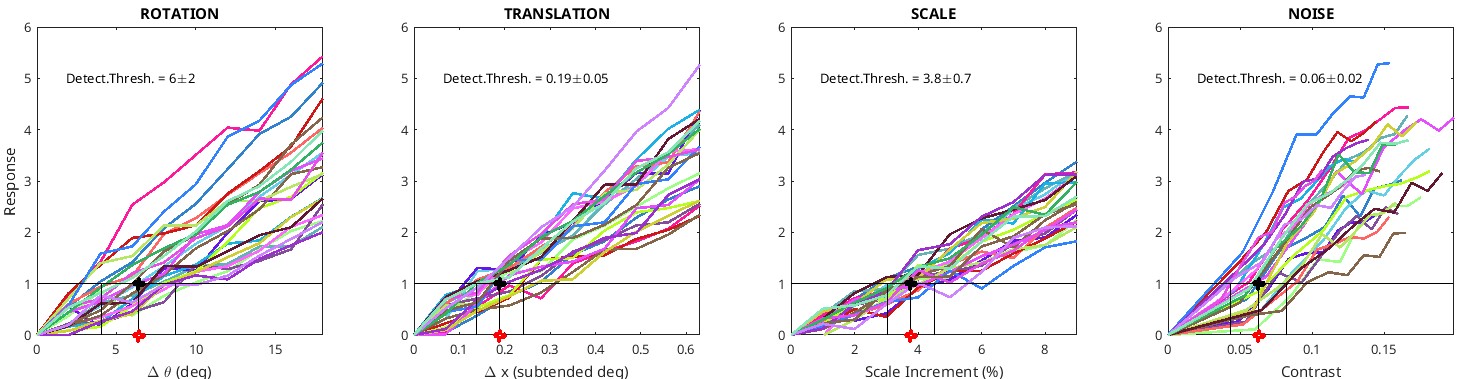}	
\caption{Absolute detection thresholds (consistent with classical literature) obtained from the inverse of the responses given the internal noise level obtained from our measurements).}
\label{fig:responses_thresh}
\end{figure}

\begin{figure}[!h]
\centering
\begin{tabular}{cc}
\includegraphics[width=0.5\linewidth]{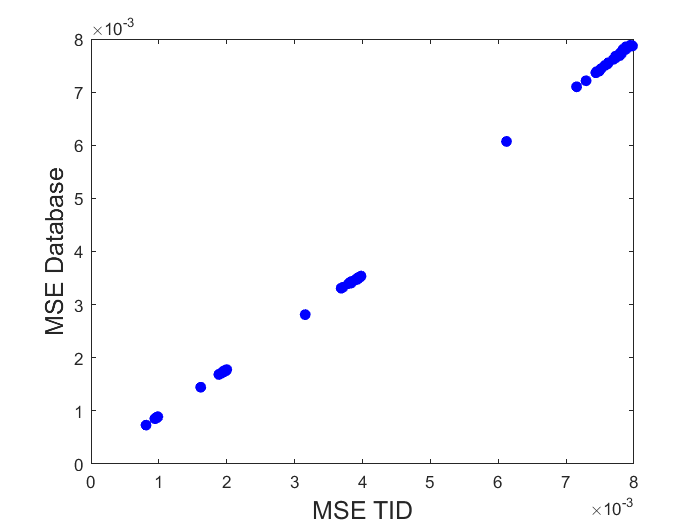}	
&			\includegraphics[width=0.5\linewidth]{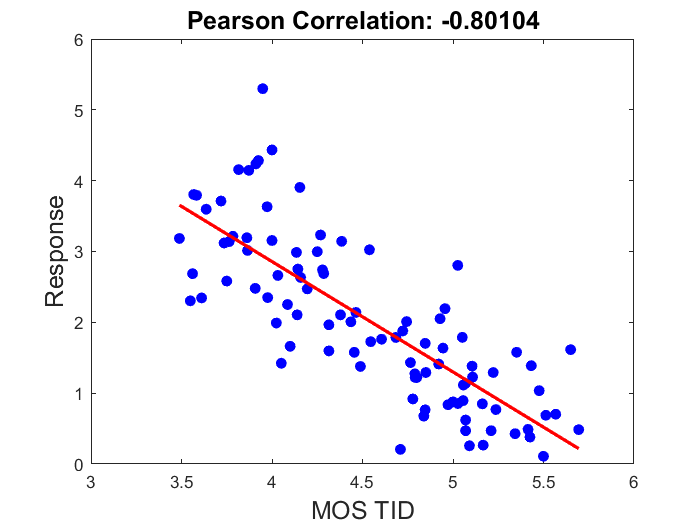}	
\end{tabular}			
\caption{Left: MSE values of distorted images with Gaussian noise, considered equivalent in our database and TID2013.
Right: Correlation between the Mean Opinion Score (MOS) in TID2013 and the responses from our database for perceptually equivalent images.}
\label{fig:tid2008}
\end{figure}

{\bf Alignment with TID and better accuracy than TID.}
We selected the highest level of Gaussian noise such that the most distorted images in our database were perceptually comparable to those in the TID2013 database. Figure \ref{fig:tid2008} presents on the left side, the MSE values for the distortions in both databases for images considered are equivalent. On the right side of the figure is presented the correlation between the responses from our database and the TID Mean Opinion Score (MOS) for these images, it demonstrates a consistent relationship between both sets of responses, being the Pearson correlation coefficient -0.8, as more distance is perceived (MLDS), less score is provided by the observer (MOS).

\begin{figure}[htbp]
    \centering
    \includegraphics[width=1\textwidth]{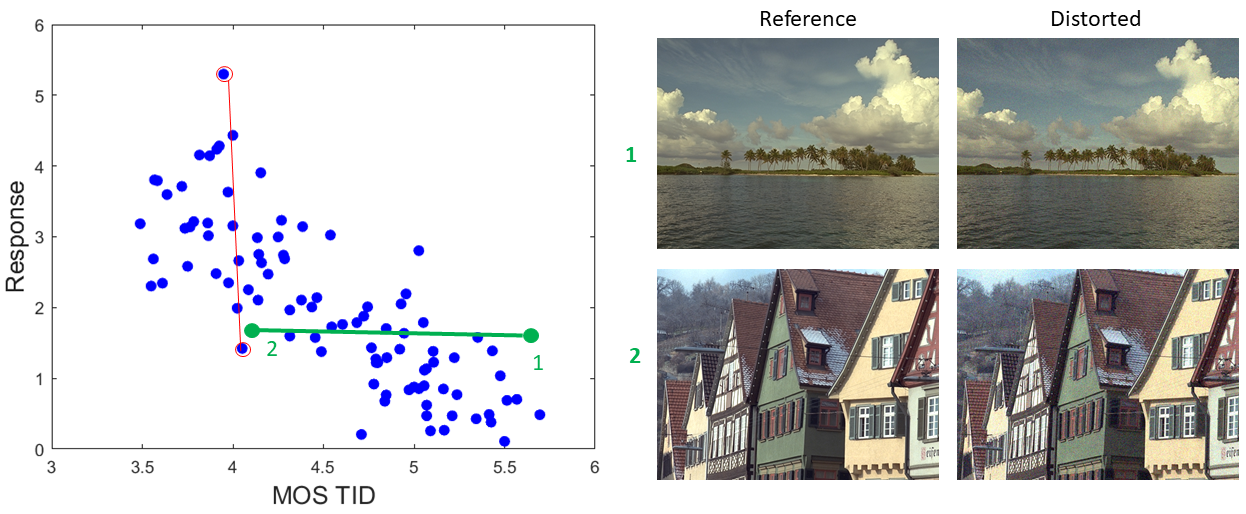}
    \includegraphics[width=1\textwidth]{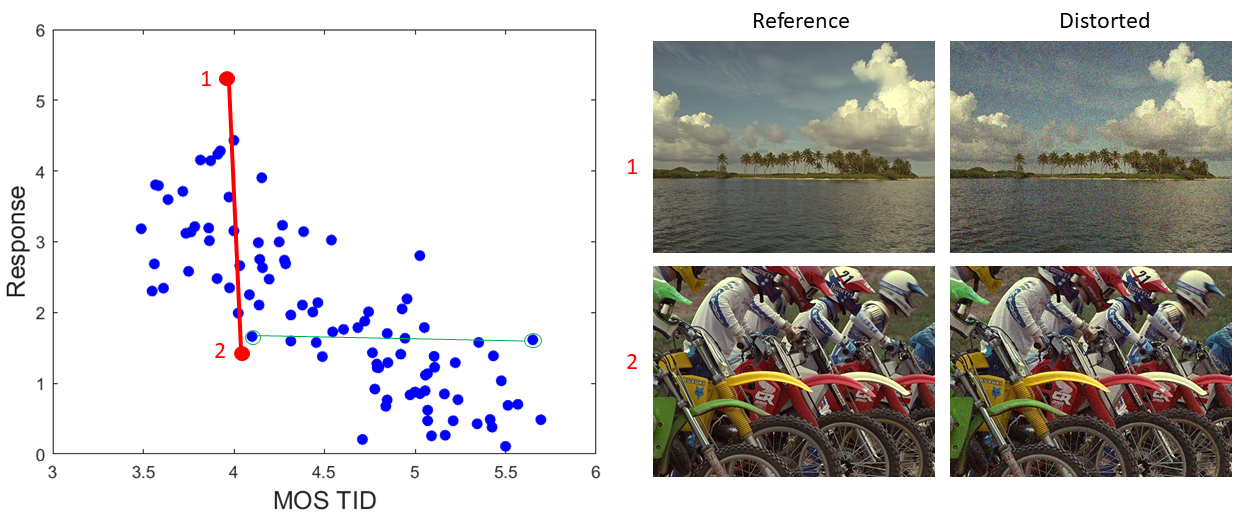}
    \caption{Group Maximum Differentiation competition. 
Top: In green, pair of distorted images and their references where our database predicts the same amount of distortion, while TID indicates that one is significantly more distorted than the other. Bottom: In red, pair of distorted images where TID predicts similar levels of distortion, but our database indicates a significant difference between them.}
    \label{fig:gmad}
\end{figure}

However, some discrepancies are observed between the responses, which may be due to the methods used to obtain them. To evaluate which measurement method gives more accurate results, we applied the group maximum differentiation (gMAD) competition \cite{gmad}. This approach compares the two sets of values by identifying pairs of images where one method predicts nearly identical responses (the distortion is perceived as the same) while the other predicts maximum perceptual differences (one distortion is much more visible than the other). Figure \ref{fig:gmad} illustrates these comparisons. In the green pair of images, our database predicts identical distortion levels, whereas the TID MOS values suggest significant differences. Visually, both images appear to have comparable distortion relative to the reference. On the other hand, the red pair of images shows cases where TID predicts similar distortion levels, but our database indicates larger differences. In this case, the top image pair exhibits a much larger perceptual difference than the bottom pair. These results suggest that our measurement method provides more reliable assessments of perceptual distortion compared to the method used in TID.

\section*{Usage Notes}



Together with the database we provide coding examples in Python. On the one hand we show how to read the databases. We also show how to read specifically the raw data and compute the MLDS responses for one image and one distortion. Other libraries can be used to compute the MLDS responses. For instance the one from the original authors \cite{JSSv025i02}, or a wrapping to use it in Python \cite{Aguilar2022}.  We also provide an example to read the MLDS responses already computed for all the images and all the distortions. The provided database can be used together with TID2008 \cite{tid2008} and TID2013 \cite{tid2013} image quality databases.

\section*{Code availability}
\label{sec:code_avail}


The code is publicly available on
GitHub under the Apache License at [\url{https://github.com/paudauo/BBDD_Affine_Transformations}]\cite{DDBBGithub}, allowing users to freely utilize and adapt it for their own research requirements. The description of each file can be found in the readme file on Github. The code has been tested in Python 3.10.12, Numpy 1.26.4, Pandas 2.2.3 and JAX 0.4.35.

\bibliography{sample}


\section*{Acknowledgements}

This work was partially funded by the Valencian local government (GVA) under the grant CIGE/2022/066 (grupos emergentes), and by the Ministerio de Ciencia e Innovación grants number PID2020-118071GB-I00 and PDC2021-121522-C21, and by the grant BBVA Foundations of Science program: Mathematics, Statistics, Computational Sciences and Artificial Intelligence (VIS4NN).

\end{document}